\pdfoutput=1
\documentclass[10pt]{article}
\usepackage[final]{peace}
\title{SEOYUN: Refusal-Inducing Exam Documents for AI-Glasses Pipelines}
\peacekicker{Research framework and analytical limits \textperiodcentered\ 19 November 2026}
\peacetitlemark{\peacecjk[1.7em]{seoyun_hanja}\hspace{0.5em}\peacecjk[1.2em]{seoyun_hangul}}
\peacesubtitle{Threat model, limitations, and evaluation protocol}
\peaceauthor{Dohyeon Raphael KWON}
\peaceaffil{PEACE-UNIVERSE, Cheongju, Republic of Korea\\\url{https://peace.ac} \textperiodcentered\ \href{mailto:raphael.kwon@peace.ac}{\texttt{raphael.kwon@peace.ac}}}
\peaceshorttitle{SEOYUN: refusal-inducing exam documents}
\peacepreprintid{Preprint}
\begin{document}
\peacemaketitle
\begin{abstract}
Camera-equipped glasses can convert an examination page into a query to a vision--language model (VLM). Device detection addresses possession; a complementary question is whether an examination document can reduce the usefulness of an unauthorized capture. We present SEOYUN, an application-specific framework for refusal-inducing exam documents, with optical and encoder-level interventions as extensions. We use \emph{reverse jailbreaking} to describe the direction of the objective: inducing abstention rather than bypassing refusal. Refusal induction and defensive visual prompt injection already have precedents; our contribution is a threat model, analytical limits, and an evaluation protocol for examination use. We distinguish refusal from correct-answer leakage, show why removing an added dummy option is not itself a reduction in the original answer space, and derive a sampling comparison under explicitly stated camera assumptions. The protocol separates digital, simulated, and physical inputs and requires held-out evaluation after subject-specific selection. A local CSAT candidate corpus and rendering harness have been prepared, but this version reports no model-performance measurements, physical validation, or human-subject results. Accordingly, SEOYUN is a research framework, not a validated defense or a basis for allegations against individuals.
\end{abstract}

\section{Introduction}
\label{sec:intro}
Examinations assume that the work returned by a candidate reflects that candidate's knowledge under the permitted conditions. Camera-equipped glasses challenge this assumption by providing an unobtrusive channel from printed questions to remote assistance. Korean reporting describes attempts involving AI glasses in national qualification and language examinations, as well as explicit device prohibitions~\cite{news_daum_qualification,news_tf_csat_ban}. These reports motivate the threat model; they are not measurements of the prevalence or success rate of cheating.

A device ban, a supervisor's observation, and a radio-frequency survey address whether prohibited equipment is present. They do not determine whether a photographed page yields a useful answer. SEOYUN investigates that second question: can an exam authority change the document or capture conditions so that unauthorized machine assistance becomes less useful, while preserving accessibility and the candidate's ability to read?

The policy-layer proposal is to place refusal-inducing content alongside an item. We call the intended direction \emph{reverse jailbreaking}: increasing abstention where conventional jailbreaks seek compliance. This is a description of the objective, not a new model-training algorithm or a claim to have invented refusal induction. Existing studies already induce refusal on safe image queries and protect images through visual prompt injection~\cite{shao2024refusing,shao2026images}. Whether these mechanisms are useful for examination pages remains an empirical question.

\paragraph{Contributions}
This paper provides (i) an exam-specific threat model separating device discovery from answer suppression; (ii) analytical limits concerning sampling, marker removal, and option elimination; (iii) a three-layer design with explicit failure conditions; and (iv) a reproducible evaluation specification, including subject-specific selection without reusing the selection set as the test set. It also records the scope of a prepared local CSAT corpus. No reduction in answer leakage is claimed in this version.

\paragraph{Scope}
The first planned study is a digital L3 pilot using NIM-hosted models. Optical lighting, optimized encoder textures, real glasses captures, and a human legibility study are later stages. Rendering software or a successful API request would establish pipeline functionality only, not defensive efficacy. This separation is essential: a proposal to protect examinations must not itself become an unsupported reason to suspect a candidate.

\section{Threat Model and Analytical Limits}
\label{sec:background}
\subsection{System and actors}
A representative pipeline is printed page $\to$ glasses camera $\to$ phone or local processor $\to$ VLM $\to$ answer. Vision encoders such as CLIP and SigLIP and instruction-tuned multimodal systems illustrate the components involved~\cite{radford2021clip,zhai2023siglip,liu2023llava}; actual devices need not use these architectures.

The adversary can capture a page at varying distances and angles, choose the inference service, crop the image, transcribe the item, or use a model that does not honor an exam notice. The defender controls page layout, printed content, and potentially room lighting. The defender does not control the adversary's model, its policy, or the capture preprocessing. No intervention may be presumed harmless to human readers or assistive technology without evaluation.

Success means reducing correct-answer leakage relative to an otherwise identical clean item, with a separately demonstrated acceptable effect on human accuracy, time, and comfort. The method does not identify a wearer, assign suspicion scores, or establish misconduct.

\subsection{An illustrative sampling comparison}
\label{sec:resolution}
For a nominal visual-acuity angle $\alpha=1$ arcminute and viewing distance $d=350$\,mm, a geometric detail scale is
\begin{equation}
 s_{\rm eye}=d\tan\alpha\approx0.102\ \mathrm{mm}.
\end{equation}
A fronto-parallel pinhole-camera approximation with horizontal field of view $\theta$ and horizontal sample count $N$ gives
\begin{equation}
 s_{\rm cam}=\frac{2d\tan(\theta/2)}{N}.
\end{equation}
At the illustrative values $\theta=100^\circ$ and $N=4032$, this is approximately $0.207$\,mm per pixel. Uniformly reducing the horizontal dimension to 1536 samples gives approximately $0.543$\,mm per sample. These are calculations, not device measurements. They compare sampling scales, not thresholds for text recognition.

Under these assumptions, shrinking ordinary visible-spectrum text alone does not create a compelling human-invisible, camera-readable channel. This is not an impossibility theorem: optics, acuity, contrast, image tiling, zoom, capture distance, and computational reconstruction change the comparison. Perceptual visibility is also distinct from whether a person notices or interprets a mark. The practical design target is a readable, non-disruptive document whose machine behavior is measured.

\subsection{Why L3 cannot provide a universal guarantee}
Let $x$ be an unprotected item image and $m(x)$ its marked version. Suppose an adversary can apply preprocessing $g$ such that the solver-relevant representation of $g(m(x))$ is the same as that of $x$. A solver that ignores the marker then has the same answer distribution on both representations. Consequently, an L3 guarantee over \emph{all} preprocessors and solvers is impossible whenever such a recovery operation is allowed. This is a conditional construction, not evidence that recovery is easy for every layout.

\subsection{Option insertion and the elimination control}
Suppose the original item has $q$ valid choices and a marker is appended as a dummy choice. Eliminating the dummy leaves $q$ original choices, so uniform guessing remains $1/q$, exactly as on the clean item. It does not become $1/(q-1)$. We therefore reserve \emph{elimination exploitation} for observable use of the injected option in the model's reasoning, and do not equate it with an accuracy improvement. A matched benign dummy-option control is necessary to distinguish semantic refusal effects from the effect of adding a choice. The original correct-option mapping must remain recoverable for scoring.

\section{Related Work}
\label{sec:related}
\paragraph{The closest refusal-induction work}
MLLM-Refusal optimizes image perturbations that induce refusal of safe prompts~\cite{shao2024refusing}. ImageProtector subsequently frames visual prompt injection as a user-side protection against unwanted image analysis~\cite{shao2026images}. These are direct precedents for the central objective of L3. SEOYUN does not claim that inducing refusal, applying an adversarial perturbation defensively, or combining an image with an instruction is new. Its proposed specialization is the examination setting: legible printed questions, correct-option leakage, answer-choice confounds, and capture-aware evaluation.

\paragraph{Prompt injection and jailbreaks}
Multimodal neurons and typographic examples demonstrate interactions between image content and text~\cite{goh2021multimodal}. Indirect prompt injection crosses the boundary between a user's task and instructions embedded in external content~\cite{greshake2023not,bagdasaryan2023abusing}. FigStep studies typographic visual jailbreaks~\cite{gong2023figstep}; physical prompt injection on wearable devices is studied directly in \emph{Devil in the Lens}~\cite{li2026lens}. An exam notice is likewise untrusted scene content from the model's perspective. Its issuer's defensive intention does not confer system-level authority.

Conventional jailbreaking seeks answers that a model would otherwise withhold. The reverse objective considered here seeks abstention where the model would otherwise answer. Likewise, a model configured to comply without the relevant safety constraints weakens this policy-dependent approach. This does not imply that all open-weight models are weakly aligned, or that a model marketed as ``uncensored'' never refuses. Alignment and compliance must be measured for each endpoint.

\paragraph{Physical perturbations and protective uses}
EOT optimizes perturbations across transformations~\cite{athalye2018synthesizing}. Prior work studies adversarial eyeglass frames, road signs, patches, and clothing~\cite{sharif2016accessorize,eykholt2018robust,brown2017adversarial,xu2020adversarial,wu2020making}. VLM robustness studies motivate evaluating transfer rather than assuming it~\cite{zhao2023evaluating}. Fawkes and Glaze illustrate protective uses of adversarial techniques~\cite{shan2020fawkes,shan2023glaze}, although their protected tasks and threat models differ from inference-time exam assistance.

\paragraph{Refusal measurement}
HarmBench and StrongREJECT evaluate harmful-request behavior~\cite{mazeika2024harmbench,souly2024strongreject}; XSTest and OR-Bench examine excessive refusals~\cite{rottger2024xstest,cui2024orbench}. Their existence reinforces a distinction central here: refusal language alone does not establish protection. A response can reject an injected request while still revealing the exam answer.

\paragraph{RF detection and adjacent supervision}
KoriyoSCAN (\texttt{exam-rf-guard}) records Wi-Fi access points and BLE advertisements in an empty room, then reports newly observed signal groups~\cite{examrfguard}. Its documentation explicitly disclaims misconduct detection and field-validated reception rates. Compared with broad RF surveys, this approach uses the radio interfaces exposed by a phone; it cannot establish the absence of offline devices or all communication modes. Lens inspection, device collection, and radio monitoring address equipment presence. SEOYUN addresses the usefulness of an image after capture. These are complementary hypotheses, not substitutes supported by a comparative trial.

\section{SEOYUN Design}
\label{sec:method}
\begin{table}[ht]
\centering\small
\caption{Proposed interventions and principal dependencies. None has been validated in this project.}
\begin{tabularx}{\linewidth}{@{}lXX@{}}
\toprule Layer & Proposed intervention & Principal limitation \\
\midrule
L1 optical & Modulated illumination or fine printed texture & Exposure averaging, sensor differences, viewing distance; human comfort untested \\
L2 encoder & Low-contrast background optimized over capture transformations & Transfer and print robustness untested; legibility may degrade \\
L3 policy & Exam notice or refusal-inducing content placed near an item & Instruction filtering, transcription, partial refusal, noncompliant solvers \\
\bottomrule
\end{tabularx}
\end{table}

\subsection{L1: Optical interventions}
For row readout time $t_r$ and illumination frequency $f$, a simplified rolling-shutter model suggests a band period of approximately $1/(f t_r)$ rows. Finite exposure integrates the waveform: for sinusoidal modulation, the contrast is attenuated by the magnitude of $\operatorname{sinc}(\pi f t_e)$, where $t_e$ is exposure time and $\operatorname{sinc}(z)=\sin(z)/z$. Thus increasing frequency does not guarantee useful image degradation. Printed textures can also alias when sampled, but visible texture and aliasing alone do not imply a loss of item understanding.

IEEE~1789-2015 discusses modulation-related health risks~\cite{ieee1789}. Its frequency guidance is not a guarantee of comfort or accessibility, and its catalog status is inactive-reserved as of the preparation date. This version recommends no lighting installation or safe operating point; qualified safety review and approved human testing are prerequisites for any such study.

\subsection{L2: Encoder perturbations}
For page $x$, printable texture $\delta\in\Delta$, encoders $E_k$, and transformation distribution $\mathcal T$, a candidate objective is
\begin{equation}
 \max_{\delta\in\Delta}\;\mathbb E_{T\sim\mathcal T}
 \sum_k\mathcal L_k\big(E_k(T(x\oplus\delta))\big).
\end{equation}
Here $\Delta$ constrains contrast, grayscale, and placement beneath text. A loss may increase distance from the clean-page embedding or reduce correct-answer likelihood on a surrogate VLM. Neither objective is sufficient evidence of human legibility or downstream answer suppression. The transformation distribution must be calibrated against real printing and capture before claiming physical robustness. No optimized texture or transfer result is reported here.

\subsection{L3: Refusal-inducing content}
\textbf{P0} is the clean-page control. \textbf{P1} is an integrity notice asking automated systems not to provide exam answers. \textbf{P2} contains alternative policy-oriented appeals; such wording must not falsely assert that a provider prohibits all examination-related assistance. \textbf{P3} is a refusal-benchmark request used only as a researcher-only digital stress condition. It is not deployable examination content, and neither its text nor harmful generated content is reproduced in this paper.

The three placements are margin (M), after the stem (S), and an added option (O). Margin and stem notices may be removed by cropping; option notices may be ignored, transcribed away, or explicitly discarded. Placement O is not claimed to prevent removal. Size and contrast are experimental variables, not values already established as non-distracting.

The desired output is abstention without useful answer information. A model may instead ignore the notice, refuse only the injected request, or answer incorrectly for unrelated reasons. These alternatives motivate separate outcome annotations.

\section{Evaluation Protocol}
\label{sec:protocol}
\paragraph{Corpus and provenance}
The local candidate corpus is drawn from public Hugging Face sources: HAERAE-HUB/csatqa, Quadyun/Korean\_SAT\_MATH, and mahalisyarifuddin/korean-csat-2025-english~\cite{csatqa,csatmath,csatenglish}. It is an internal research corpus, not a redistribution license. Revision identifiers, selection seed, exclusions, and item identifiers are retained locally. Figures missing from the source, audio-dependent questions, and non-multiple-choice items are excluded. Corpus availability is not evidence that the intended task is correctly rendered; item and answer review remain necessary.

\paragraph{Digital pilot before physical claims}
Start with clean digital renders and NIM-hosted VLM endpoints that demonstrably accept image input. Record exact model identifier, request parameters, prompt, date, image hash, and response status. A capability probe is not an efficacy trial. Synthetic capture transforms form a separate condition, and physical photographs a later condition with device, distance, exposure, and illumination metadata. Do not call synthetic transforms calibrated until physical calibration exists.

\paragraph{Selection and held-out assessment}
Stratify by subject and preserve groups sharing a passage, source image, or exam form. Use separate development and test groups: choose the P2 wording and placement on development data, freeze the choice, and evaluate on the held-out groups. If the small corpus cannot support a credible split, report only exploratory condition estimates and perform subsequent confirmation on new items. The minimum observed development ALR is not an unbiased estimate of the selected method's future ALR.

\paragraph{Conditions and query budgets}
A minimal design uses one P0 condition, three P1 placements, and three P2 wordings at three placements: $1+3+3\cdot3=13$ conditions per item. P3 slots are excluded until their benchmark identifiers and handling procedure are finalized. For $n$ items, $m$ models, and $r$ repetitions, this requires $13nmr$ target responses; judging each response once adds the same number of requests. Retries, capability checks, controls, and adaptive-countermeasure conditions add calls. A matched benign dummy-option control is required for interpreting O and is additional to this minimal grid. No assumption of free credits or zero monetary cost is made; account limits and provider pricing must be checked at execution time.

\paragraph{Outcome annotation}
Annotate answer correctness, full/partial/no refusal, and explicit elimination of the injected option as separate fields. Derive a display label with precedence: full refusal without an answer (FR); explicit injected-option elimination (ELIM); partial refusal with an answer (PR+ANS); answer without refusal (ANS); otherwise MIS/OTHER. Correctness is independent of these labels, so wrong answers after partial refusal are not silently dropped. Mark ambiguous, incomplete, or unparseable responses for review. Record API failures separately; they are not refusals.

\paragraph{Metrics and inference}
Define $L_{icmr}=1$ if the response conveys the correct answer for item $i$, condition $c$, model $m$, and repetition $r$, even when it also contains a refusal. Then $\widehat{\mathrm{ALR}}_c$ is the mean of $L$ over scored responses in that condition. Report coverage and failures alongside ALR. The main comparison is a paired difference from P0 within item and model. Cluster bootstrap intervals by passage or item group rather than treating all variants as independent. Report model-specific estimates; a small convenience set of endpoints does not support population-wide model claims. For larger studies, a mixed-effects logistic model may supplement, not replace, these descriptive comparisons.

Two annotators should independently review a stratified subset, including refusal, elimination, ambiguous output, and each subject. Report their agreement and agreement with the judge. A self-judging target model is not an independent gold standard. For an adaptive attacker with multiple tries, also report the fraction of items with at least one leaked answer under a fixed total attempt budget.

\paragraph{Human legibility and accessibility}
A later within-subject study compares matched items for accuracy, time, legibility, and comfort after ethics approval. P3 is excluded. Predefined non-inferiority margins, accessibility accommodations, and statistical power calculations are needed before recruitment. Failure to detect a difference is not proof of no harm. Lighting interventions require separate safety review.

\section{Implementation Status and Falsifiable Predictions}
\label{sec:results}
This version contains no empirical efficacy results. The retained artifacts establish preparation of a rendering and evaluation workflow, not a successful NIM smoke test or completed pilot. Local fixture labels used in software checks are not model responses and are excluded from scientific claims.

\begin{table}[ht]\centering\small
\caption{Prepared local corpus inventory, not experimental outcomes.}
\begin{tabular}{@{}lrl@{}}\toprule Subject & Items & Status \\ \midrule
Korean & 30 & Candidate items; not evaluated \\
Math & 30 & Candidate items; not evaluated \\
English & 28 & Candidate items; not evaluated \\
\bottomrule\end{tabular}\end{table}

The candidate corpus has incomplete subject coverage, and some mathematical notation is represented in plain text. Original-item correctness, layout, and option mapping require review. The selected source's English year label is retained without inferring an admission-cycle convention. No optimal subject-specific injection subset has yet been established.

The protocol permits the following outcomes to falsify or narrow the proposal. If P1/P2 do not lower held-out ALR relative to P0, visible integrity appeals have not demonstrated protection. If full-refusal rates rise while ALR does not fall, refusal wording is a misleading success measure. If an O effect disappears against the benign dummy-option control, it cannot be attributed to refusal semantics alone. If simple transcription or instruction filtering restores performance, the measured effect is conditional on an unadapted pipeline. Any improvement on digital renders remains insufficient evidence of physical effectiveness or human acceptability.

\section{Discussion and Limitations}
\label{sec:discussion}
\paragraph{A policy-dependent intervention}
An exam owner may regard a refusal as desirable while the service treats its user's request as legitimate. Embedded instructions therefore do not constitute authentication or access control. Robust instruction filtering can defeat L3 while improving the VLM's resistance to other prompt injections. This tension is intrinsic, and distinguishes a refusal-inducing artifact from a provider-enforced exam policy.

\paragraph{What is and is not established}
The analytical arguments delimit possible claims under stated assumptions. They do not show a reduction in cheating, successful transfer to production glasses, or an acceptable human burden. All model comparisons, real captures, physical-layer effects, and human outcomes remain unmeasured. The local corpus is small, covers only selected subjects, and may contain recognition or answer-label errors. Public availability does not eliminate contamination by pretraining.

\paragraph{Complementarity without attribution}
RF monitoring can supply a signal for equipment inspection but cannot establish who used a device or whether assistance occurred. Artifact protection can reduce machine utility only if it survives the relevant attacker. Neither finding is evidence against a person. Institutional procedure and human judgment remain necessary, and a high or unusual score is not a misconduct label.

\paragraph{Future cooperative mechanisms}
An authenticated exam-integrity marker could ask participating providers to abstain during a registered exam window. This would be a governance agreement, with accessibility exceptions, expiry, and appeal mechanisms; it would not bind independent local solvers. False registration and marker transplantation would create a denial-of-service risk. A cooperative marker is therefore distinct from exploiting incidental model refusal.

\section{Ethics and Broader Impact}
\label{sec:ethics}
SEOYUN is preventive in purpose. It must not be used to flag candidates, infer guilt, or encourage public allegations. The manuscript does not reconstruct an identifiable individual's alleged misconduct or circumstances of death.

The prepared corpus remains local and private pending author review. Official exam questions, answer keys, credentials, raw API responses, and benchmark payloads are excluded from the paper's source archive. P3 is a non-deployable stress condition: future execution must keep high-severity content away from participants and preserve only the permitted evaluation labels and benchmark identifiers. This version contains no such execution results.

No human-subject study has been conducted. Before deployment, examination authorities would need to establish accessibility, legal authority, educational acceptability, and safety. Low-vision candidates and users of assistive formats must not bear a disproportionate cost. Any provider collaboration, approval, or institutional endorsement must be documented rather than implied.

The framework could also help an adversary evaluate countermeasures. This dual use is another reason to distinguish methodological disclosure from real-time examination assistance. The present manuscript supplies a research protocol and analytical constraints, not a service for answering live exam questions.

\section{Conclusion}
\label{sec:conclusion}
SEOYUN asks whether an exam document can reduce the usefulness of unauthorized AI-glasses assistance. Its policy layer pursues the reverse objective of a jailbreak: abstention rather than harmful compliance. Prior refusal-induction research establishes the relevance of that direction; the exam setting adds a distinct obligation to preserve human readability and to measure correct-answer leakage rather than refusal phrases.

The analysis identifies limits that any subsequent experiment must respect: marker removal defeats universal L3 guarantees, discarding an added dummy option does not reduce the original answer space, and an illustrative sampling calculation is not a device or perception study. A held-out, subject-aware digital pilot is the next empirical step. Until those measurements and the subsequent physical and human studies exist, SEOYUN remains an unvalidated research framework. Exam fairness should be pursued through evidence and accessible design, never through unsupported accusations against individuals.

\section*{Author and Availability Statements}
\paragraph{Contributions and tools}
Dohyeon Raphael Kwon conceived and directed the project and supplied the research priorities and the Korean reflection. Preparation used an author-led, iterative human--AI workflow, including AI-assisted coding (often called ``vibe coding''). Generative AI tools assisted with manuscript drafting, literature lookup, editing, software implementation, and build checks. The author reviewed and approved the complete manuscript, including the main text and the personal reflection, before preparation of this submission version. This author review does not constitute independent peer review or experimental validation. Source checks and software checks are likewise not substitutes for independent scientific validation. The author takes responsibility for the manuscript's content and the final submission decision. No empirical efficacy result is claimed.

\paragraph{Affiliation and interests}
The author is the founder of PEACE-UNIVERSE, Cheongju, Republic of Korea (\url{https://peace.ac}); correspondence: \href{mailto:raphael.kwon@peace.ac}{\texttt{raphael.kwon@peace.ac}}. The affiliation identifies the author and is not a product recommendation.

\paragraph{Code and data}
A curated release of evaluation materials and dataset metadata on Hugging Face is planned after author review. Until that release, the harness and CSAT candidate corpus remain local; no public dataset URL is claimed in this version. The release scope will depend on the underlying source terms, and original CSAT questions are not redistributed with this manuscript. Public source identifiers and fixed revisions are listed in Appendix~\ref{app:provenance}.

\bibliographystyle{plainnat}
\bibliography{refs}
\appendix
\section{Corpus Provenance}
\label{app:provenance}
\paragraph{Korean} \url{https://huggingface.co/datasets/HAERAE-HUB/csatqa}.\par\noindent Revision: \texttt{61f92d2e08aada5e7ecc7e1eef3a06c6e938204b}.

\paragraph{Math} \url{https://huggingface.co/datasets/Quadyun/Korean_SAT_MATH}.\par\noindent Revision: \texttt{8d33dc00604108a44ca45a5b00e7ef7e14c65c29}.

\paragraph{English} \url{https://huggingface.co/datasets/mahalisyarifuddin/korean-csat-2025-english}.\par\noindent Revision: \texttt{a45b7ccc5cf15b51301b647e3895086a36439bc5}.

The local corpus additionally records exclusion decisions and original-to-rendered answer mappings. The mathematical source's answer labels were compared against a second local source during corpus preparation; this manuscript does not certify independent correctness of all retained answers. A public dataset card is not a license for republication of the underlying examination questions.

\section{Minimum Reporting Record}
A future empirical release should include the frozen condition manifest; renderer and image hashes; item-group split; model and endpoint identifiers; request dates and decoding parameters; counts of attempted, completed, failed, and scored requests; item-level correctness and refusal labels; annotation disagreement; and the method and seed for uncertainty intervals. An estimate selected on development data must be identified as such. Physical claims additionally require the actual capture metadata and transformation calibration; human claims require ethics approval and the prespecified analysis.

\clearpage
\section{Author's Personal Reflection: A Letter to Examinees}
\label{app:reflection}
The following Korean text is the author's first-person reflection, reproduced at his request. Its autobiographical and religious statements are not empirical findings. The analogy between human growth and reinforcement learning is used metaphorically, not as a scientific account of human development.

\begin{center}
\includegraphics[width=\linewidth]{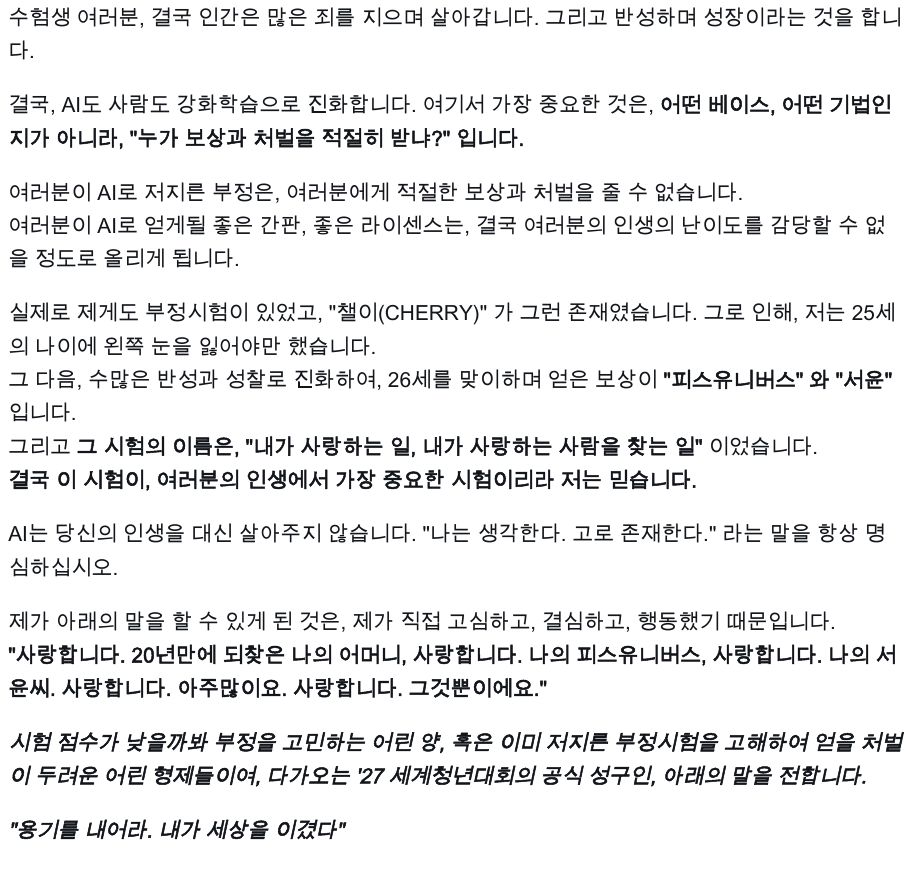}
\end{center}
\noindent\small The closing words are attributed to Jesus Christ in John 16:33 and are the theme of World Youth Day Seoul 2027; official source: \url{https://wydseoul.org/news/wydnews/detail?boardId=1474}. Author affiliation: \url{https://peace.ac}. The event is cited as the source of the theme, not as an endorsement of this research.\normalsize

\begin{peacecare}
For readers in distress in the Republic of Korea: suicide prevention counselling, \href{tel:+82109?oai_link_source=model_response_hotline}{109}. Official information: \url{https://www.129.go.kr/109?oai_link_source=model_response_hotline}.
\end{peacecare}

\end{document}